\documentclass[nohyperref]{article}

\usepackage{microtype}
\usepackage{graphicx}
\usepackage{booktabs} 

\usepackage{hyperref}

\usepackage[accepted]{icml2022}

\usepackage{amsmath}
\usepackage{amssymb}
\usepackage{mathtools}
\usepackage{amsthm}

\usepackage[capitalize,noabbrev]{cleveref}

\theoremstyle{plain}

\theoremstyle{definition}

\theoremstyle{remark}

\usepackage[textsize=tiny]{todonotes}

\usepackage{multirow}
\usepackage{graphicx}
\usepackage{subfig}
\usepackage{dblfloatfix}
\usepackage{enumitem}

\def\ie{\emph{i.e.}}
\def\etal{\emph{et al.}}

\icmltitlerunning{Gradient-Based Adversarial and Out-of-Distribution Detection}

\begin{document}
\thispagestyle{empty}
\twocolumn[{%
\pagestyle{empty}
\vspace{30mm}
{ \large
\begin{itemize}[leftmargin=2.5cm, align=parleft, labelsep=2cm, itemsep=4ex,]

\item[\textbf{Citation}]{J. Lee, M. Prabhushankar, and G. AlRegib, “Gradient-Based Adversarial and Out-of-Distribution Detection,” in International Conference on Machine Learning (ICML) Workshop on New Frontiers in Adversarial Machine Learning, Baltimore, MD, USA, July 2022.
}

\item[\textbf{Review}]{Date of Acceptance: June 13, 2022}


\item[\textbf{Bib}]  {@inproceedings\{lee2022adversarial,\\
    title=\{Gradient-Based Adversarial and Out-of-Distribution Detection\},\\
    author=\{Lee, Jinsol, and Prabhushankar, Mohit, and AlRegib, Ghassan\},\\
    booktitle=\{International Conference on Machine Learning (ICML) Workshop on New Frontiers in Adversarial Machine Learning\},\\
    year=\{2022\}\}}

\item[\textbf{Contact}]{
\{jinsol.lee, alregib\}@gatech.edu\\
\url{https://ghassanalregib.info/}\\}
\end{itemize}
}}]
\newpage
\clearpage

\twocolumn[
\icmltitle{Gradient-Based Adversarial and Out-of-Distribution Detection}



\icmlsetsymbol{equal}{*}

\begin{icmlauthorlist}
\icmlauthor{Jinsol Lee}{gt}
\icmlauthor{Mohit Prabhushankar}{gt}
\icmlauthor{Ghassan Alregib}{gt}
\end{icmlauthorlist}

\icmlaffiliation{gt}{Omni Lab for Intelligent Visual Engineering and Science (OLIVES), School of Electrical and Computer Engineering, Georgia Institute of Technology, Atlanta, GA 30332}

\icmlcorrespondingauthor{Jinsol Lee}{jinsol.lee@gatech.edu}

\icmlkeywords{Machine Learning, ICML}

\vskip 0.3in
]



\printAffiliationsAndNotice{}  

\begin{abstract}
We propose to utilize gradients for detecting adversarial and out-of-distribution samples.
We introduce \textit{confounding labels}---labels that differ from normal labels seen during training---in gradient generation to probe the \textit{effective expressivity} of neural networks.
Gradients depict the amount of change required for a model to properly represent given inputs, providing insight into the representational power of the model established by network architectural properties as well as training data.
By introducing a label of different design, we remove the dependency on ground truth labels for gradient generation during inference.
We show that our gradient-based approach allows for capturing the anomaly in inputs based on the \textit{effective expressivity} of the models with no hyperparameter tuning or additional processing, and outperforms state-of-the-art methods for adversarial and out-of-distribution detection.
\end{abstract}

\section{Introduction}

Deep neural networks (DNNs) have achieved significant advancements over the past decade, and they are increasingly utilized in cybersecurity applications such as malware detection and intrusion detection~\cite{mahdavifar2019application}.
However, many recent studies pointed out that DNNs are prone to failure when deployed in real-world environments where they often encounter data that diverge from training conditions~\cite{temel2017curetsr, temel2018cureor, temel2019multifarious}.
Neural networks rely heavily on the implicit assumption that any given input during inference is drawn from the same distribution as the training data (\ie, in-distribution), thereby making overconfident predictions~\cite{guo2017calibration}.
In addition, the discovery of adversarial attacks by Szegedy~\etal~\cite{szegedy2013intriguing} highlighted the vulnerability of DNNs to inputs with small perturbations specifically designed to deceive the models.
To ensure reliable performance of practical applications of a neural network, the model must be capable of distinguishing inputs that differ from training data and cannot be handled properly.

\begin{figure}[t]
\centering
\includegraphics[width=0.99\linewidth]{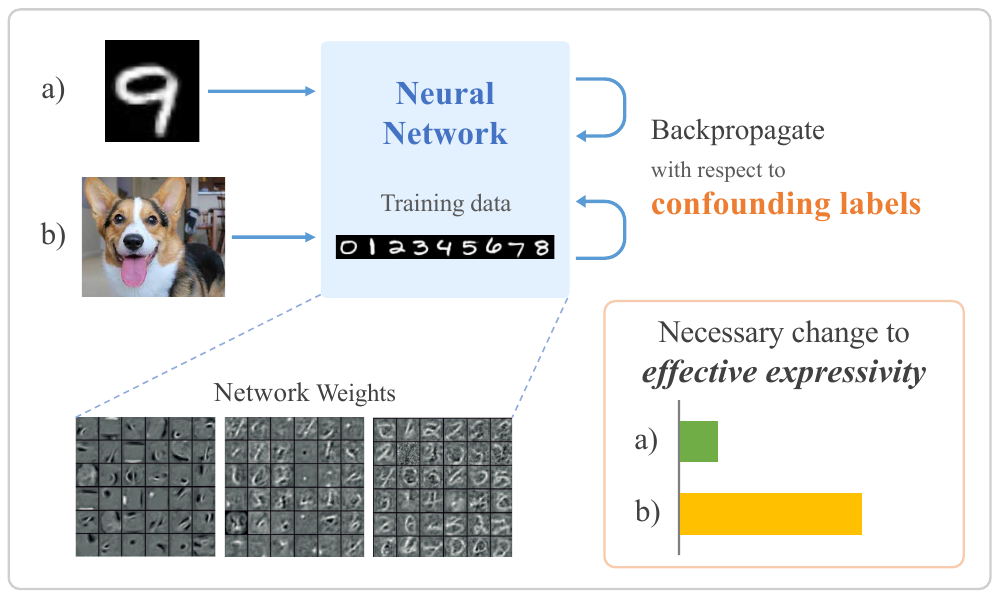}
\caption{Overview of our approach. We argue that the necessary change observed in gradients invoked by a \textit{confounding label} is smaller for an input similar to training data (\ie, image \textbf{a}) than a more anomalous input from the model's perspective (\ie, image \textbf{b}).}
\label{fig:confounding_label}
\end{figure}

In this paper, we examine the ability of trained neural networks to handle inputs based on its \textit{effective expressivity}.
We define \textit{effective expressivity} as the representational power of a model based on the training data.
We propose to utilize backpropagated gradients to analyze the \textit{effective expressivity} of a trained model.
Gradients correspond to the amount of changes that a model requires to accurately represent a given sample.
We argue that with gradients, we can characterize the anomaly in inputs based on what the model is unfamiliar with and thus incapable of representing properly.
We describe the motivation of our approach in Fig.~\ref{fig:confounding_label}.
A classifier is trained with handwritten digit images of 0 through 8 to capture their characteristics with its weights, visualized for some layers of the trained classifier in Fig.~\ref{fig:confounding_label}.
During inference, we observe the necessary changes captured in gradients invoked by a \textit{confounding label} with respect to an image similar to training data (\ie, a handwritten digit 9) and highly dissimilar (\ie, dog).
Our hypothesis is that the necessary change to \textit{effective expressivity} would be larger for the dissimilar input since the model weights learned on digit 0 to 8 would not be able to capture the characteristics of a dog.
One problem in utilizing gradients to observe the necessary change is that we do not have access to the labels for given inputs or any information regarding their distribution.
To remove dependency on ground truth labels in gradient generation during inference, we introduce a \textit{confounding label}---label that the network has not seen during training.
While many studies proposed to manipulate the design of labels for various purposes~\cite{tokozume2018between, zhang2017mixup, yun2019cutmix, durand2019learning, duarte2021plm}, they only analyze the effect of different label designs in training with the statistics of training dataset.
In comparison, a \textit{confounding label} does not require the knowledge of training data and can be used during inference with no access to ground truth labels.
Our methodology allows for capturing anomaly in inputs from the perspective the model based on its \textit{effective expressivity}.
The contributions of this paper are three-fold:
\vspace{-1mm}
\begin{itemize}[leftmargin=5mm,itemsep=0.4mm]
  \item We propose to utilize backpropagated gradients to characterize anomaly in inputs seen during inference from the perspective of the model.
  \item We introduce \textit{confounding labels} as a tool to elicit model response that can be utilized to probe the \textit{effective expressivity} of trained neural networks.
  \item We validate our approach on the applications of detecting anomalous inputs including adversarial and out-of-distribution (OOD) samples, and achieve state-of-the-art performance with no hyperparameter tuning or additional processing.
\end{itemize}
\vspace{-2mm}

\section{Related Work and Preliminary}

\subsection{Adversarial \& Out-of-Distribution Detection}

There are many recent studies of detecting anomalous samples, whether they are adversarially generated or due to statistical shift in data distribution.
\citet{hendrycks2016baseline} introduced a baseline method to threshold samples based on the predicted softmax distributions.
\citet{liang2018odin} proposed additional input and output processing to further improve the baseline method. 
\citet{devries2018confidence} utilized prediction confidence obtained from an augmented confidence estimation branch on a pre-trained classifier. 
\citet{ma2018lid} proposed to characterize the dimensional properties of adversarial regions with local intrinsic dimensionality. 
\citet{lee2018mahalanobis} proposed a confidence metric using Mahalanobis distance. 
\citet{liu2020energyOOD} utilized energy score to capture the likelihood of occurrence during inference and training time. 
Most approaches utilize activation-based representations to characterize inputs via learned features, \ie, what the network knows about given inputs.
However, for anomalous inputs, the overconfident nature of neural networks~\cite{guo2017calibration} makes it counterintuitive to characterize them solely with the learned features.
Rather, we argue that the anomaly in inputs should be established from the perspective of the model based on what the model is unfamiliar with and thus incapable of representing properly.

\begin{figure*}[t]
\centering
\vspace{-2mm}
\subfloat[Gradient-based representation generation]{
    \label{sfig:gradients}
    \includegraphics[height=5.5cm]{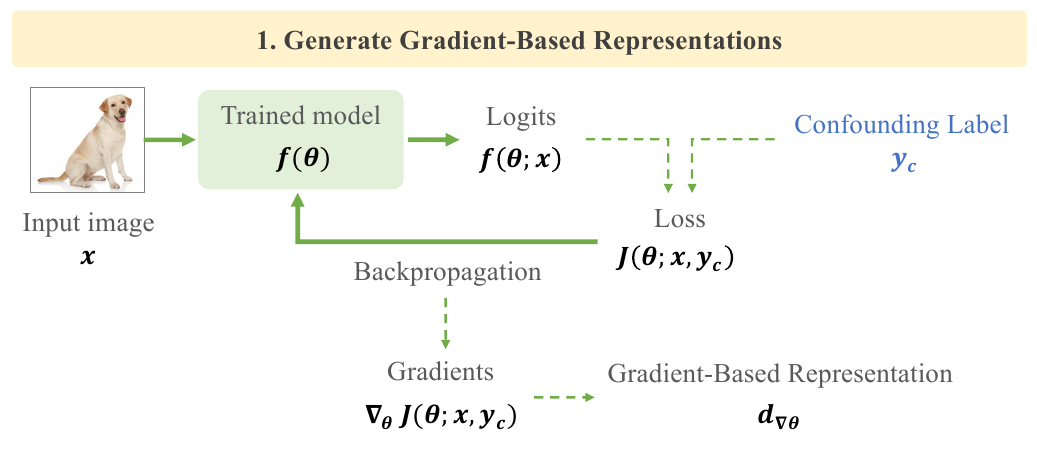}}
\hspace{4mm}\vrule  \hspace{2mm}
\subfloat[Detection]{\label{sfig:detection}
    \includegraphics[height=5.5cm]{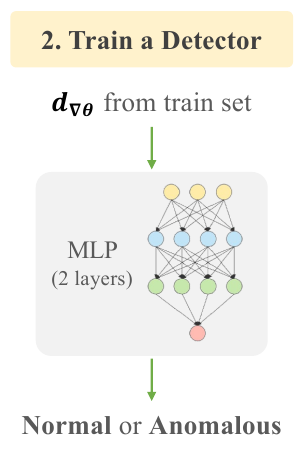}}
\caption{Overall framework for gradient-based anomalous input detection.}
\label{fig:framework}
\end{figure*}

\subsection{Gradients as Features}

At the core of training neural networks lies gradient-based optimization techniques~\cite{ruder2016gradient}. 
Apart from their utility as a tool to search for a converged solution, backpropagated gradients have been utilized for various purposes.
We discuss the recent studies based on how the relevant gradients are generated.
First category is by backpropagating model outputs (\ie, logits) directly.
This is widely observed in visualization techniques~\cite{selvaraju2017gradcam, chattopadhay2018grad} to highlight pixels based on classification scores.
\citet{zinkevich2017holographic} also utilizes logits to extract holographic representation of the network.
These approaches assume that the test data is drawn from the same distribution as the training data, and they observe the model response by further enforcing the output of the model.
Another category of approaches is by backpropagating loss between logits and some target labels.
Gradient-based adversarial attack generation fits under this category.
\citet{goodfellow2014adversarial} was first to demonstrate that a small but intentional perturbation can affect models to completely change their predictions with high confidence.
Others proposed adversarial attacks that also utilize gradients in an iterative manner~\cite{kurakin2017bim_iterll, madry2017pgd} with different initialization constraints. 
Another relevant work include contrastive visual explanation~\cite{prabhushankar2020contrastive} where model classification is contrasted with other possible classes.
\citet{oberdiek2018classification, lee2020gradients,lee2021osr} also utilize loss-based approaches to generate gradients to quantify the uncertainty of neural networks.
\citet{kwon2019distorted, kwon2020backpropagated} employ the directional information of gradients for representation learning.
These approaches utilize gradients in a similar way to model training by incorporating some forms of class labels seen during training.
It is natural to utilize these in-distribution class labels when the considered test data is safe to be assumed in-distribution and it allows for inciting the model response relevant for the classes of interest.
However, this becomes irrelevant when the data during inference is not drawn from the same distribution as the training data.
We argue that to characterize anomaly in inputs, \textit{unseen} labels should be introduced for gradient generation instead of seen class labels.

\subsection{Manipulating Label Designs}


One-hot encoding is the most widely used approach in formulating labels to handle nominal data (\ie, categorical data without quantitative relationships between categories) for many applications including single-label image classification.
For multi-label image classification, a combination of multiple one-hot encoded vectors is used to indicate information about multiple classes in each image. 
Based on these two, many recent studies analyze different formulations of labels for various purposes.
Some work~\cite{tokozume2018between, zhang2017mixup, yun2019cutmix} proposed to mix two input images and their labels with pre-computed ratio as a data augmentation technique to improve generalizability as well as robustness of the models.
\citet{prabhushankar2021extracting} explored combinations of binary classification labels in extracting causal visual features for interpretability.
Another type of manipulated label design is partial labels.
\citet{durand2019learning} tackled the problem of missing labels per image in multi-label classification and proposed to exploit the proportion of known labels per sample to weigh the contribution of the images and labels.
\citet{duarte2021plm} proposed to mask parts of labels based on the ratio of positive samples to negative samples to handle imbalanced dataset.
These approaches explore different designs of labels only in model training and require the knowledge of training data.
We propose to manipulate label designs to characterize inputs from the perspective of models with gradients during inference.
Contrary to existing studies, our approach does not depend on the availability of training data statistics or ground truth labels of inference data.

\subsection{Capacity and Expressivity}

We discuss the ability of a trained neural network to handle given inputs based on two main concepts: \textit{capacity} and \textit{expressivity}.
\textit{Capacity} of a neural network is mainly expressed by its size or the number of parameters~\cite{neyshabur2017exploring, sun2017revisiting}.
On the other hand, \textit{expressivity} describes how different architectural properties of a network (depth, width, layer type) affects its ability to approximate functions~\cite{raghu2017expressive, zhang2017understanding}.
Many studies discuss \textit{expressivity} by analyzing different classes of neural networks with the same number of parameters on the functions which the networks are capable or incapable of representing.
In essence, \textit{capacity} can be seen as the maximum utilization of the network's representational power whereas \textit{expressivity} is the observed utilization based on the connection of network parameters. 
\section{Method}

In this section, we discuss our framework in two stages as shown in Fig.~\ref{fig:framework}: generating gradient-based representations with \textit{confounding labels} and detecting anomalous inputs.
We also demonstrate the effectiveness of gradients generated with \textit{confounding labels} in comparison with activations. 

\subsection{Gradients with Confounding Labels}\label{ssec:gradients_cf_labels}

\begin{figure*}[t]
\centering
\vspace{2mm}
\subfloat{
    \includegraphics[width=0.60\textwidth]{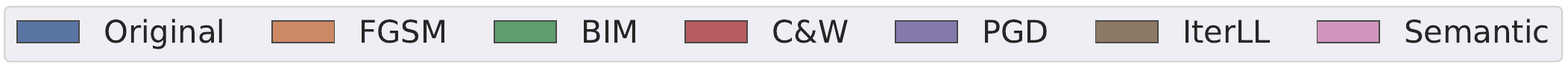}
   }\vspace{-2mm}\hfill
\setcounter{subfigure}{0}%
\subfloat[Gradients]{\label{sfig:mag_grad}
    \includegraphics[width=0.49\textwidth]{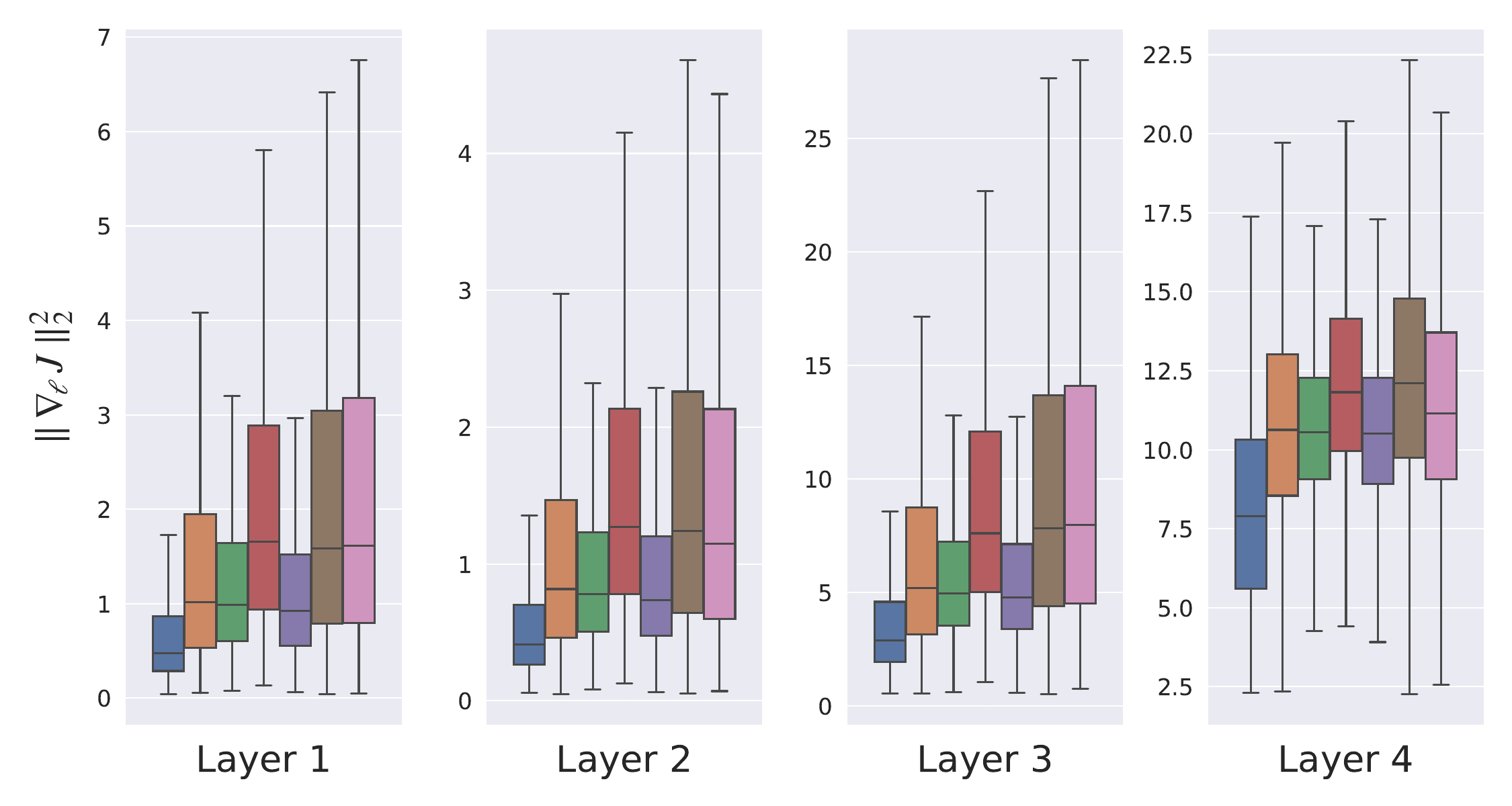}
    }
\subfloat[Activations]{\label{sfig:mag_activ}
    \includegraphics[width=0.49\textwidth]{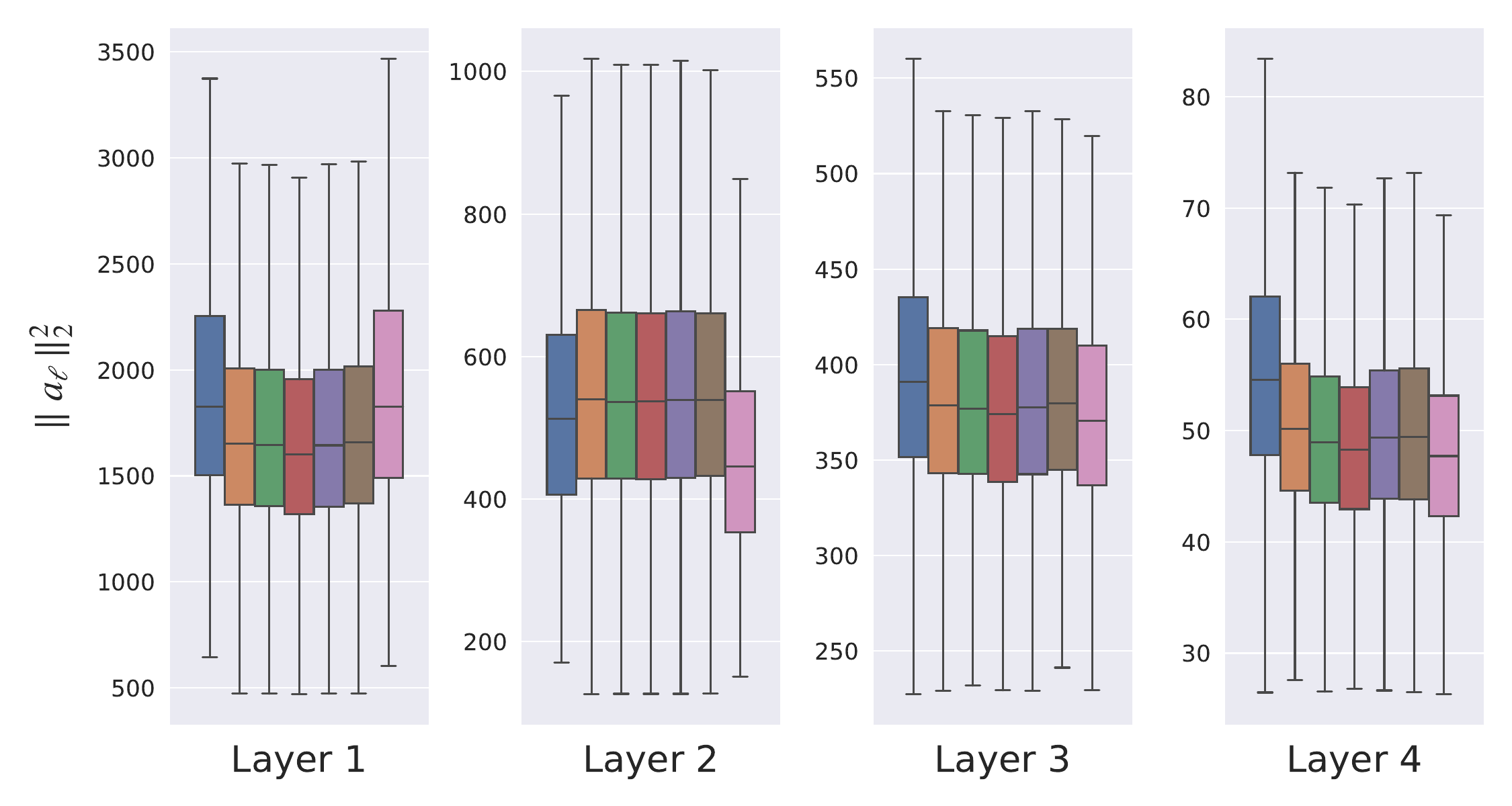}
    }
\caption{Comparison between layer-wise gradients $\nabla_\ell J$ and activations in $L2$-norms. The magnitude values are obtained from a convolutional layer from each residual block of a ResNet-18 model trained on CIFAR-10 dataset. }
\label{fig:grad_vs_activ}
\end{figure*}

Based on the concepts of \textit{capacity} and \textit{expressivity}, we define \textit{effective expressivity} as the \textit{expressivity} that is effectively utilized for learning given training data, which establishes the knowledge base of the model.
Consider two models of the same architecture but with different training datasets of MNIST~\cite{lecun1998mnist} and ImageNet~\cite{deng2009imagenet}.
The \textit{capacity} and \textit{expressivity} would be the same for both models since the network architecture remains fixed and these characteristics are independent of data.
However, the \textit{effective expressivity} would be less for the model trained with a simpler dataset of MNIST than more complex dataset of ImageNet.
We propose to utilize gradients generated with a \textit{confounding label} to examine the \textit{effective expressivity} of a classifier during inference.
Our approach focuses on the lack of knowledge that the model should offset in order to fully grasp the anomaly in the data. 
A \textit{confounding label} provides an unsupervised way to elicit gradient responses relevant for the \textit{effective expressivity} of a trained network that is necessary to improve its representational power and get closer to its \textit{capacity}.

We discuss the first stage of our framework, which is the generation of gradient-based representations with \textit{confounding labels}, shown in Fig.~\ref{sfig:gradients}.
An input image is passed through a trained image classifier, and the model yields logits.
We utilize binary cross entropy loss  $J(\theta)$ between the logits and a \textit{confounding label}, as shown in Eq.~(\ref{eq:bce}), where $\hat{y_i}$ is the predicted probability for class $i$, $y_{c,i}$ is the true probability represented by the \textit{confounding label}, and $N$ is the number of classes.
For this paper, we utilize the \textit{confounding label} that combines one-hot encodings of all classes (\ie, a vector of all 1's).
\begin{equation} \label{eq:bce}
    J(\theta)= \frac{1}{N} \sum^{N-1}_{i=0} \left(y_{c,i} \cdot \log\left(\hat{y_i}\right) + \left(1 - y_{c,i}\right)\cdot\log\left(1-\hat{y_i}\right)\right),
\end{equation}
The loss is then backpropagated through the model, generating gradients for each network layer.
While any form of the gradients that preserve their magnitude would be valid, we measure the squared $L_{2}$-norm of the gradients for each model parameter set (\ie, weight and bias parameters of network layers) and concatenate them to construct a gradient-based representation for each input.
The obtained representation is of the following form,
\begin{equation}
  \begin{gathered}
    d_{\nabla \theta} = \big[~\|\nabla J_{\theta_0}(\theta)\|^2_2~,~\cdots~,~\|\nabla J_{\theta_{L-1}}(\theta)\|^2_2~\big],
  \end{gathered}
\end{equation}
where $L$ is the number of layers in a given network.

\vspace{2mm}
\noindent\textbf{Validation of gradients with \textit{confounding labels}}~~
We demonstrate the effectiveness of gradients obtained with \textit{confounding labels} in comparison with activations.
In an adversarial detection setup, we show the disparity between \textit{effective expressivity} established with training data and the necessary adjustments to represent original image samples and adversarial samples.
The key idea is that the model will need larger updates to handle adversarial samples than original images, captured by the magnitudes of the generated gradients.
For this experiment, we first train a ResNet-18~\cite{he2016resnet} classifier trained with original CIFAR-10~\cite{krizhevsky2009cifar} train set.
The test set of CIFAR-10 is utilized to generate adversarial attacks of the following: fast gradient sign method (FGSM)~\cite{goodfellow2014adversarial}, basic iterative method (BIM)~\cite{kurakin2017bim_iterll}, Carlini \& Wagner attack (C\&W)~\cite{carlini2017cw}, projected gradient descent (PGD)~\cite{madry2017pgd}, iterative least-likely class method (IterLL)~\cite{kurakin2017bim_iterll}, and semantic attack~\cite{hosseini2017semantic}.
We collect the L2 norm of layer-wise gradients and activations and visualize their distribution in Fig.~\ref{fig:grad_vs_activ}.
Due to the space limitation, we select a convolutional layer from each residual block of the ResNet architecture for visualization.
The separation in the ranges of gradient magnitudes, shown in Fig.~\ref{sfig:mag_grad}, between original images of CIFAR-10 test set and their adversarial versions is more evident in some parts of the network than others because each layer captures information about different aspects of given inputs.
But overall, we observe the smaller magnitude ranges for original images and significantly larger magnitude ranges for adversarial attack images.
On the other hand, the activations throughout the network, shown in Fig.~\ref{sfig:mag_activ}, do not show as much variation in the magnitude ranges for different image sets.
It supports our intuition that gradients can characterize the anomaly better than activations based on the unfamiliar aspects of given inputs from the model's perspective.

\begin{table*}[ht]
\captionsetup{width=.9\linewidth}
\caption{Adversarial detection results for CIFAR-10 in AUROC. For Mahalanobis method (denoted as M), we report vanilla results (V), \ie, without the input pre-processing or feature ensemble, input pre-processing only (P), feature ensemble only (FE), and with both (P+FE). All values are in percentages and the best results are highlighted in bold.}
\label{table:adversarial}
\vskip 0.1in
\begin{center}
\begin{small}
\begin{sc}
\resizebox{0.85\linewidth}{!}{
\begin{tabular}{clccccccc}
\toprule
Model & Attacks & Baseline & LID & M(V) & M(P) & M(FE) & M(P+FE) & Ours \\
\midrule
\multirow{6}{*}{ResNet} & FGSM & 51.20 & 90.06 & 81.69 & 84.25 & \textbf{99.95} & \textbf{99.95} & 93.45 \\
 & BIM & 49.94 & 99.21 & 87.09 & 89.20 & \textbf{100.0} & \textbf{100.0} & 96.19 \\
 & C\&W & 53.40 & 76.47 & 74.51 & 75.71 & 92.78 & 92.79 & \textbf{97.07} \\
 & PGD & 50.03 & 67.48 & 56.27 & 57.57 & 65.23 & 75.98 & \textbf{95.82} \\
 & IterLL & 60.40 & 85.17 & 62.32 & 64.10 & 85.10 & 92.10 & \textbf{98.17} \\
 & Semantic & 52.29 & 86.25 & 64.18 & 65.79 & 83.95 & 84.38 & \textbf{90.15} \\
 \midrule
\multirow{6}{*}{DenseNet} & FGSM & 52.76 & 98.23 & 86.88 & 87.24 & \textbf{99.98} & 99.97 & 96.83 \\
 & BIM & 49.67 & \textbf{100.0} & 89.19 & 89.17 & \textbf{100.0} & \textbf{100.0} & 96.85 \\
 & C\&W & 54.53 & 80.58 & 75.77 & 76.16 & 90.83 & 90.76 & \textbf{97.05} \\
 & PGD & 49.87 & 83.01 & 70.39 & 66.52 & 86.94 & 83.61 & \textbf{96.77} \\
 & IterLL & 55.43 & 83.16 & 70.17 & 66.61 & 83.20 & 77.84 & \textbf{98.53} \\
 & Semantic & 53.54 & 81.41 & 62.16 & 62.15 & 67.98 & 67.29 & \textbf{89.55} \\
\bottomrule
\end{tabular}}
\end{sc}
\end{small}
\end{center}
\end{table*}

\subsection{Anomalous Input Detection}\label{ssec:detection}

We discuss the second stage of our framework in Fig.~\ref{sfig:detection}, which is the detection of anomalous inputs with gradient-based representations. 
The gradient-based representations are collected for both normal and anomalous images with \textit{confounding labels}, and they are split into train, validation, and test sets (40\%-40\%-20\% split).
Each representation is labeled with the binary information regarding its corresponding input image (positive for anomalous inputs, negative for normal inputs).
A multi-layer perceptrons (MLP) of 2 layers is trained as an anomalous input detector using the train and validation sets, and the trained MLP model is applied on the test set to report the detection accuracy.
 
\section{Experiments}

In this section, we utilize gradient-based representations generated in response to \textit{confounding labels} for detecting anomalous inputs: adversarial detection and out-of-distribution detection.

\subsection{Adversarial Detection}


For adversarial detection, we train ResNet and DenseNet~\cite{huang2017densenet} architectures for a classification task on CIFAR-10 training set. 
The test set of CIFAR-10 is utilized to generate adversarial attacks mentioned in the validation portion of Section~\ref{ssec:gradients_cf_labels}. 
We utilize the pristine CIFAR-10 test set as negative samples and the adversarial images as positive samples. 
For comparison, we employ the Baseline method~\cite{hendrycks2016baseline}, local intrinsic dimensionality (LID) scores~\cite{ma2018lid}, and the Mahalanobis method~\cite{lee2018mahalanobis}. 
Following the setup detailed in Section~\ref{ssec:detection}, the performance is measured in AUROC and reported in Table~\ref{table:adversarial}. 
For the Mahalanobis method, we report vanilla results (V), \ie, without input pre-processing or feature ensemble, results with input pre-processing only (P), results with feature ensemble only (FE), results with both (P+FE) to demonstrate its dependency on delicate calibration of hyperparameters.

The Mahalanobis method shows the best performance for the adversarial attack types of FGSM and BIM. 
On the other hand, our approach outperforms all state-of-the-art methods for the C\&W, PGD, IterLL and Semantic attacks. 
Note that the Mahalanobis method requires fine-tuning of hyperparameters. 
For all attacks, we utilize the range of hyperparameter values that they have explored in their paper and report the performance with the best parameter values.
Interestingly enough, the Mahalanobis approach shows less saturated performance for the types of attacks that were not included in their work. 
These attack types also correspond to the ones for which our method outperforms all the state-of-the-art approaches.
This indicates the need for further hyper parameter tuning outside the suggested parameter values, depending on the types of considered adversarial attacks.
It highlights the benefit of our approach in its simplicity of obtaining the gradient-based representations with no hyperparameter tuning or any pre-/post-processing. 
We remark that our method can characterize the adversarial attack samples effectively, even for the attack types that are widely considered to be highly challenging to detect: C\&W and PGD attacks.

\subsection{Out-of-distribution Detection}

\begin{table*}[t]
\captionsetup{width=.95\linewidth}
\caption{Out-of-distribution detection results in detection accuracy, AUROC, and AUPR. All values are in percentages and the best results are highlighted in bold.}
\label{table:ood}
\begin{center}
\begin{small}
\begin{sc}
\resizebox{\linewidth}{!}{
\begin{tabular}{clccc}
\toprule
\multicolumn{2}{l@{}}{Dataset Distribution} & \multicolumn{3}{c@{}}{Baseline / ODIN / Mahalanobis / Energy / Ours} \\
\cmidrule(l){3-5}
 In  &  Out & Detection Accuracy & AUROC & AUPR \\
\midrule
CIFAR-10        & SVHN & 83.36 / 88.81 / 91.95 / - / \textbf{98.04} & 88.30 / 94.93 / 97.10 / 95.72 / \textbf{99.84} & 88.26 / 95.45 / 96.12 / 97.26 / \textbf{99.98} \\
    (ResNet)        & ImageNet & 84.01 / 85.21 / \textbf{97.45} / - / 91.28 & 90.06 / 91.86 / \textbf{99.68} / 84.95 / 96.92 & 89.26 / 91.60 / \textbf{99.60} / 84.60 / 97.21 \\       
                    & LSUN & 87.34 / 88.42 / \textbf{98.60} / - / 98.37 & 92.79 / 94.48 / \textbf{99.86} / 92.39 / \textbf{99.86} & 92.30 / 94.22 / 99.82 / 95.40 / \textbf{99.87} \\ \midrule
    SVHN            & CIFAR-10 & 79.98 / 80.12 / 88.84 / - / \textbf{97.90} & 81.50 / 81.49 / 95.05 / 98.85 / \textbf{99.79} & 81.01 / 80.95 / 90.25 / \textbf{99.57} / 98.11 \\
    (ResNet)        & ImageNet & 81.70 / 81.92 / 96.17 / - / \textbf{97.74} & 83.69 / 83.82 / 99.23 / 98.18 / \textbf{99.77} & 82.54 / 82.60 / 98.17 / \textbf{99.30} / 97.93 \\       
                    & LSUN & 80.96 / 81.15 / 97.50 / - / \textbf{99.04} & 82.85 / 82.98 / 99.54 / 97.72 / \textbf{99.93} & 81.97 / 82.01 / 98.84 / 99.09 / \textbf{99.21} \\ \midrule
    CIFAR-10        & SVHN & 87.71 / 86.80 / 95.75 / - / \textbf{98.92} & 93.51 / 94.51 / 98.96 / 89.33 / \textbf{99.94} & 94.60 / 95.20 / 97.21 / 94.01 / \textbf{99.99} \\
    (DenseNet)      & ImageNet & 84.93 / 91.16 / \textbf{96.83} / - / 90.07 & 91.50 / 96.93 / \textbf{99.45} / 86.17 / 96.44 & 91.02 / 96.85 / \textbf{99.29} / 87.20 / 96.41 \\       
                    & LSUN & 85.58 / 91.91 / 98.08 / - / \textbf{99.20} & 92.09 / 97.51 / 99.74 / 92.77 / \textbf{99.96} & 91.71 / 97.51 / 99.72 / 95.57 / \textbf{99.96} \\ \midrule
    SVHN            & CIFAR-10 & 86.61 / 86.25 / 96.50 / - / \textbf{98.34} & 91.90 / 91.76 / 99.02 / 90.63 / \textbf{99.82} & 92.12 /  92.38 /  94.40 / 94.30 / \textbf{98.36} \\
    (DenseNet)      & ImageNet & 90.22 / 90.31 / \textbf{98.89} / - / 98.21 & 94.78 / 95.08 / \textbf{99.84} / 87.51 / 99.81 & 94.76 / 95.53 / \textbf{99.37} / 92.15 / 98.16 \\
                    & LSUN & 89.14 / 89.13 / 99.49 / - / \textbf{99.52} & 94.12 / 94.52 / 99.92 / 83.45 / \textbf{99.96} & 94.32 / 95.20 / 99.54 / 89.33 / \textbf{99.70} \\
\bottomrule
\end{tabular}}
\end{sc}
\end{small}
\end{center}
\end{table*}

We employ various image classification datasets for OOD detection: CIFAR-10 and SVHN~\cite{netzer2011svhn} as in-distribution, and additional TinyImageNet~\cite{deng2009imagenet} and LSUN~\cite{yu2015lsun} as OOD. 
We demonstrate our approaches using ResNet-18 and DenseNet. 
For evaluation, we measure detection accuracy, area under receiver operating characteristic curve (AUROC), and area under precision-recall curve (AUPR). 
The results of OOD detection are reported in Table~\ref{table:ood} with other state-of-the-art methods, including Baseline~\cite{hendrycks2016baseline}, ODIN~\cite{liang2018odin}, Mahalanobis~\cite{lee2018mahalanobis}, and Energy~\cite{liu2020energyOOD}. 
For Energy method, detection accuracy is omitted as they do not specifically determine the threshold values for energy scores for OOD detection.

We observe that our method outperforms other state-of-the-art methods in AUROC when CIFAR-10 is used as in-distribution and SVHN as OOD, and SVHN as in-distribution and all others as OOD for ResNet. 
For DenseNet, our method performs better in every experiment except with TinyImageNet as OOD. 
Our method is especially effective when there is a larger difference in the complexity of in-distribution and OOD datasets. 
The results support our intuition behind the \textit{effective expressivity} of a trained network based on its training data.
The models trained on CIFAR-10 have better \textit{effective expressivity} than those trained on the simpler SVHN, and they are capable of extracting more diverse features. 
The models trained on SVHN would require larger updates to handle inputs of higher complexity, exhibiting more apparent gap in the model responses and resulting in the better performance in OOD detection.

\begin{figure}[h]
\centering
\subfloat[CIFAR-10]{\label{sfig:dataset_cifar10}
    \includegraphics[width=0.20\linewidth]{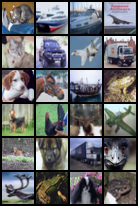}
  }\hfill
\subfloat[SVHN]{\label{sfig:dataset_svhn}
    \includegraphics[width=0.20\linewidth]{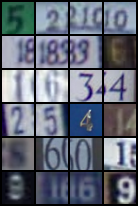}
  }\hfill
\subfloat[ImageNet]{\label{sfig:dataset_tinyimagenet}
    \includegraphics[width=0.20\linewidth]{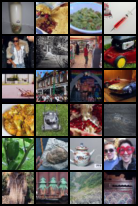}
  }\hfill
\subfloat[LSUN]{\label{sfig:dataset_lsun}
    \includegraphics[width=0.20\linewidth]{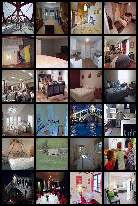}
  }
\caption{Sample dataset images.}\label{fig:dataset_samples}
\end{figure}

It is interesting to note the performance of our gradient-based approach when TinyImageNet is used as OOD for both architectures. 
With SVHN as the in-distribution dataset and TinyImageNet as OOD, our approach achieves the best performance, if not a very close second. 
However, with CIFAR-10 as in-distribution, our approach shows inferior performance to the Mahalanobis method. 
We provide samples images of each dataset for visual analysis in Fig.~\ref{fig:dataset_samples}. 
With CIFAR-10 as in-distibution, TinyImageNet shows the most visual similarity with CIFAR-10 in the types and scales of objects in the images among the three OOD datasets shown in Fig.~\ref{fig:dataset_samples}~\subref{sfig:dataset_svhn}-\subref{sfig:dataset_lsun}. 
SVHN is a street-view house number dataset, which is vastly different from CIFAR-10 of natural images. 
LSUN is a scene understanding dataset with scene categories as classifications.
Our experiment results show that the more similar the in-distribution and OOD datasets, the smaller the gap in the spreads of gradient magnitudes, which makes it more challenging to differentiate OOD samples from in-distribution samples. 
Overall, we remark that our gradient-based method outperforms all activation-based methods in most cases.

\vspace{-1mm}
 \vspace{-3mm}
\section{Conclusion}
In this paper, we propose to examine the \textit{effective expressivity} of neural networks with backpropagated gradients for adversarial and out-of-distribution detection.
With gradients, we characterize the anomaly in inputs seen during inference based on the lack of knowledge of the model, rather than with its learned features.
We introduce \textit{confounding labels} as a tool to elicit gradient response to remove the dependency on ground truth labels during inference.
We show that our approach outperforms state-of-the-art methods for adversarial and out-of-distribution detection with no need for hyperparameter tuning or additional processing.

\bibliography{bib}
\bibliographystyle{icml2022}

\end{document}